%% file: root.tex
\documentclass[letterpaper, 10 pt, journal, twoside]{IEEEtran}  % Comment this line out if you need a4paper

\usepackage{graphics} % for pdf, bitmapped graphics files
\usepackage{float}
\usepackage[colorlinks = true,
            linkcolor = blue,
            urlcolor  = blue,
            citecolor = blue,
            anchorcolor = blue]{hyperref}
            
\usepackage{balance}
\usepackage{xcolor}

\usepackage[disable]{todonotes}

\graphicspath{{graphics/}}

\usepackage{subcaption}
\newcommand{\revision}[1]{#1}

\begin{document}
% Paper headers
\markboth{IEEE Robotics and Automation Letters. Preprint Version. Accepted November, 2020}
{Kratzer \MakeLowercase{\textit{et al.}}: MoGaze: A Dataset of Full-Body Motions that Includes Workspace Geometry and Eye-Gaze} 
% Use only for final RAL version

%\author{Philipp Kratzer$^{1}$, Simon Bihlmaier$^2$, Niteesh Balachandra Midlagajni$^2$, Rohit Prakash$^2$,\\ Marc Toussaint$^{3}$ and Jim Mainprice$^{1}$\\% <-this % stops a space
%\vspace{0.1cm}
%\authorblockA{$^1$\tt{\small{firstname.lastname@ipvs.uni-stuttgart.de}}}
%\authorblockA{$^1$Machine Learning and Robotics Lab, University of Stuttgart, Germany}
%\authorblockA{$^2$Humans to Robots Motions Research Group ; HRM ; University of Stuttgart, Germany}
%\authorblockA{$^3$Learning and Intelligent Systems Lab ;  Technical University of Berlin, Germany}
%\vspace{-0.8cm}
%}

% Make room for more info lines in the \author command 
\author{Philipp Kratzer$^{1}$, Simon Bihlmaier$^2$, Niteesh Balachandra Midlagajni$^2$, Rohit Prakash$^2$,\\ Marc Toussaint$^{3}$ and Jim Mainprice$^{1}$
\thanks{Manuscript received: July, 1, 2020; Revised September, 18, 2020; Accepted November, 8, 2020.}%Use only for final RAL version
\thanks{This paper was recommended for publication by Editor Tamim Asfour upon evaluation of the Associate Editor and Reviewers' comments.
This work was supported by the research alliance ``System Mensch'' funded by the German Federal Ministry for Science, Research and Arts.} %Use only for final RAL version
\thanks{$^{1}$Philipp Kratzer and Jim Mainprice are with the Machine Learning and Robotics Lab, University of Stuttgart, Germany and the Humans to Robots Motions Research Group, University of Stuttgart, Germany
        {\tt\footnotesize philipp.kratzer@ipvs.uni-stuttgart.de; jim.mainprice@ipvs.uni-stuttgart.de}}%
\thanks{$^{2} $ Simon Bihlmaier, Niteesh Balachandra Midlagajni and Rohit Prakash are with the Humans to Robots Motions Research Group, University of Stuttgart, Germany
  {\tt\footnotesize simon.bihlmaier@gmail.com; midlagajni@psychologie.tu-darmstadt.de; rohitt.747@gmail.com}}%
\thanks{$^{3} $ Marc Toussaint is with the Learning and Intelligent Systems lab, TU Berlin, Germany
        {\tt\footnotesize toussaint@tu-berlin.de}}
\thanks{Digital Object Identifier (DOI): see top of this page.}
}
% Use only for final RAL version.

\title{MoGaze: A Dataset of Full-Body Motions that Includes Workspace Geometry and Eye-Gaze}

\maketitle
\begin{figure*}
  \newcommand{\ltscale}{.245}
  \includegraphics[width=\ltscale\linewidth]{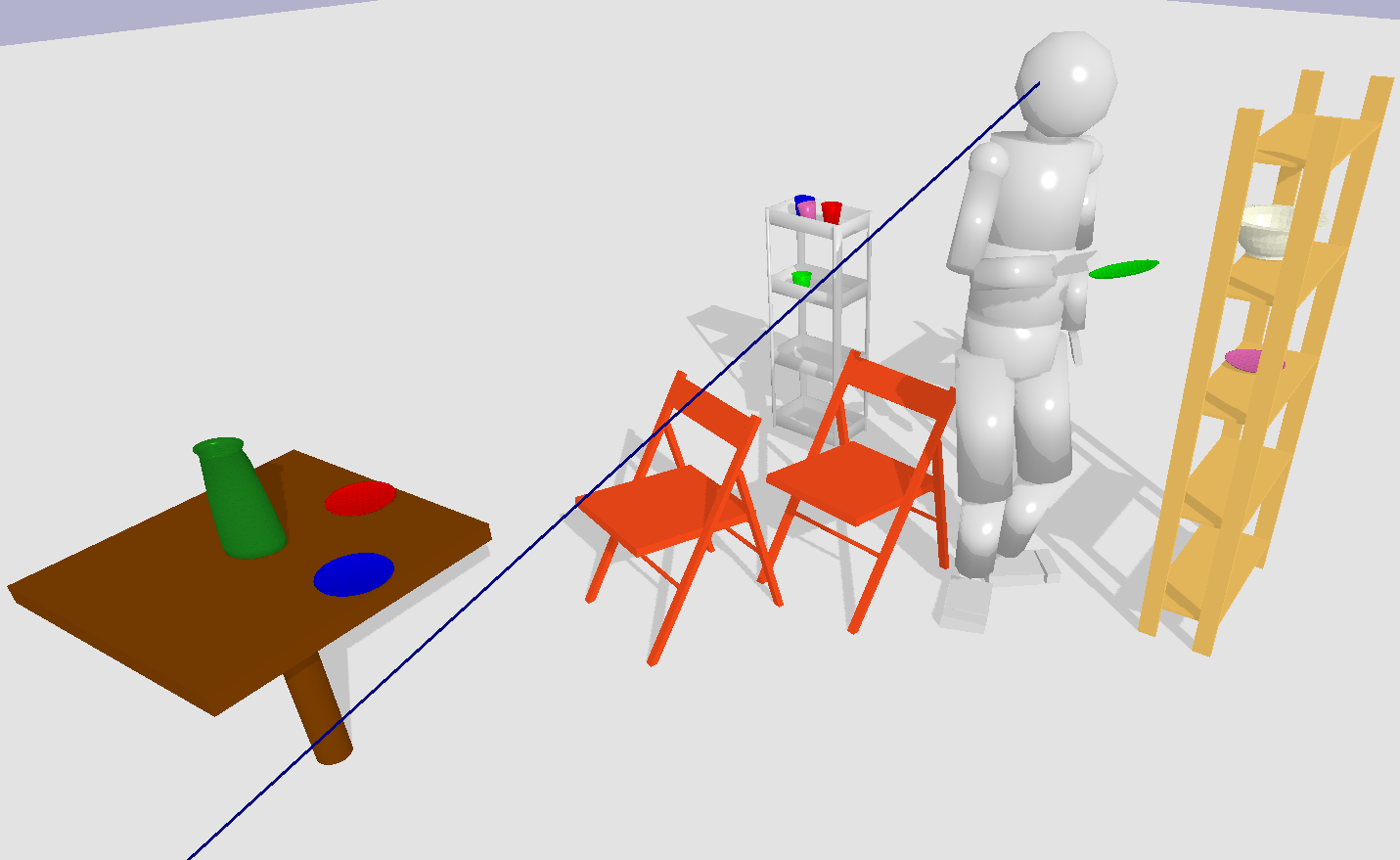}
  \includegraphics[width=\ltscale\linewidth]{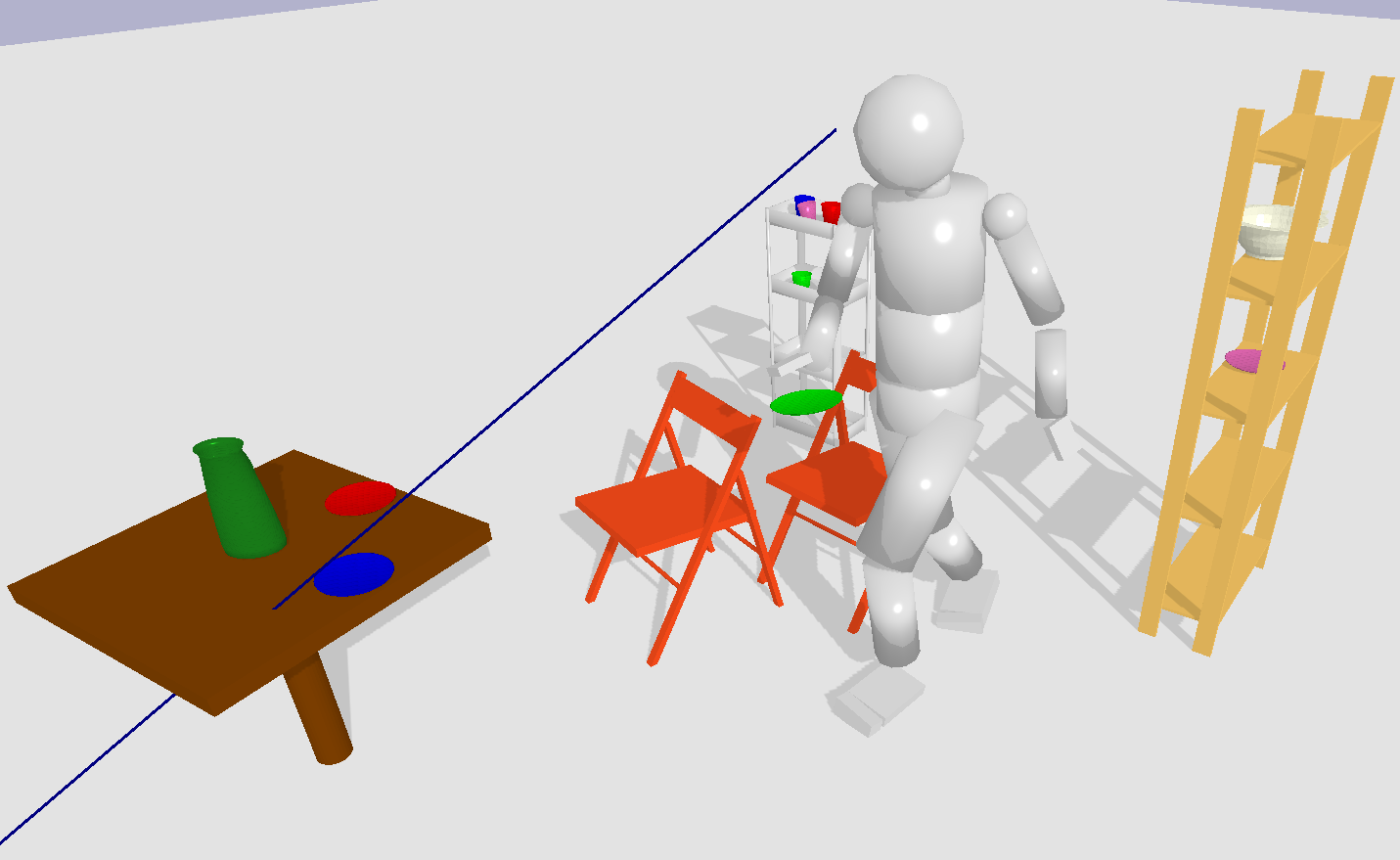}
  \includegraphics[width=\ltscale\linewidth]{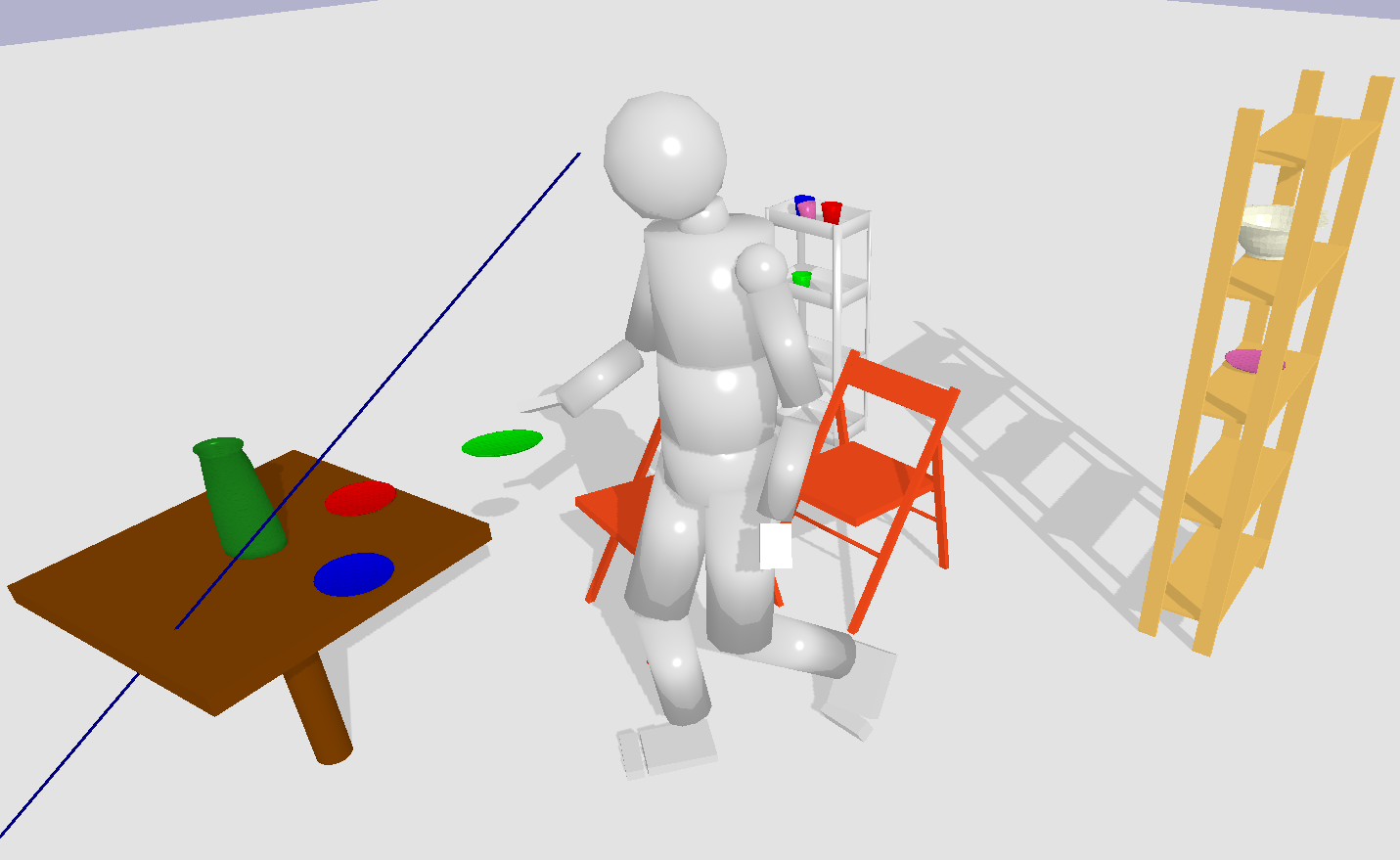}
  \includegraphics[width=\ltscale\linewidth]{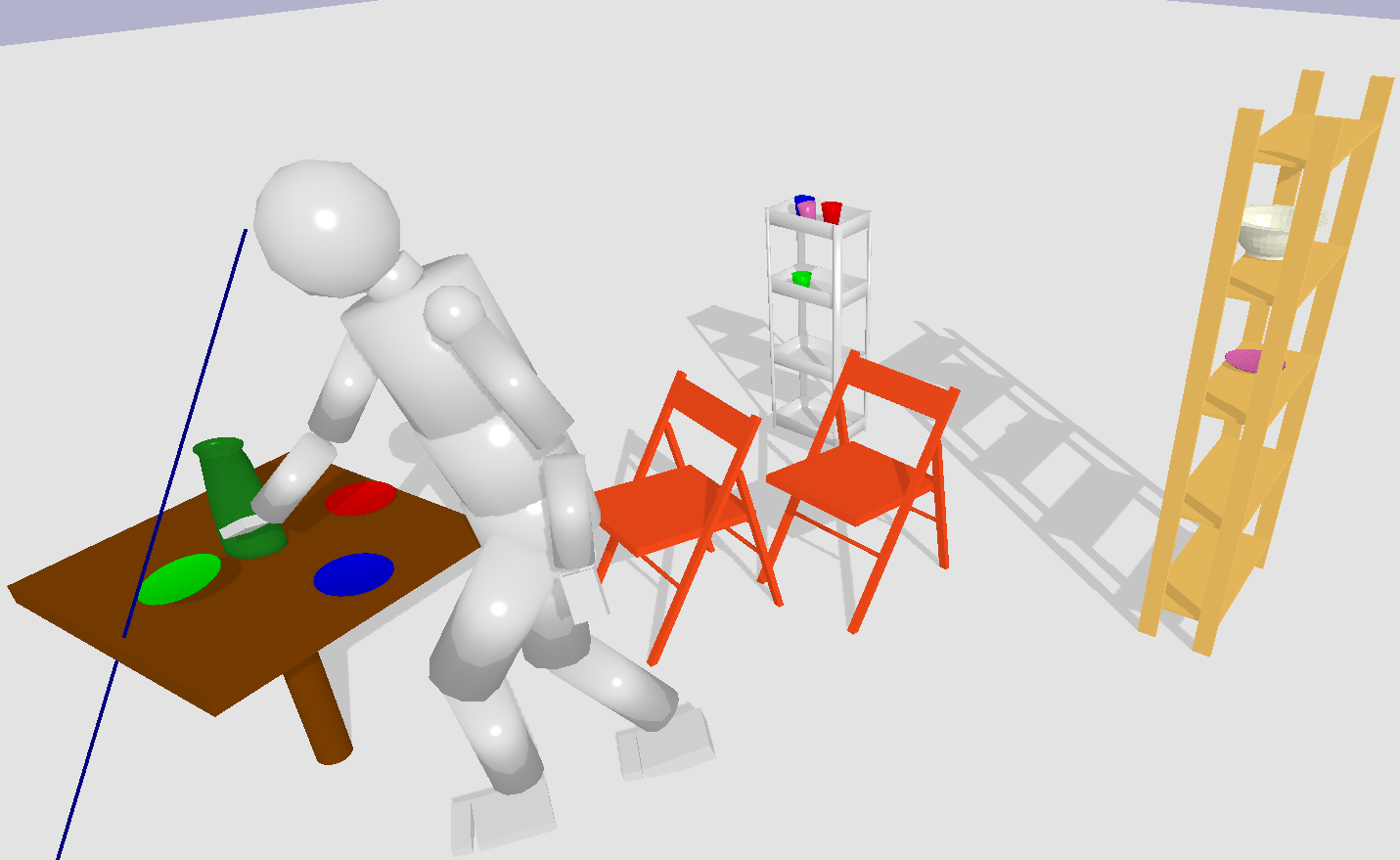}
  \caption{A human grasped a plate from the shelf and places it on the table. It can be observed how the gaze shifts towards the actual placing position even before the human reaches the table.}
  \label{fig:place_plate}
\end{figure*}

%\twocolumn[{%
%\renewcommand\twocolumn[1][]{#1}%
%\maketitle
%\begin{center}
%  \centering
%  \newcommand{\ltscale}{.245}
%  \includegraphics[width=\ltscale\linewidth]{gazetraj/im1}
%  \includegraphics[width=\ltscale\linewidth]{gazetraj/im2}
%  \includegraphics[width=\ltscale\linewidth]{gazetraj/im3}
%  \includegraphics[width=\ltscale\linewidth]{gazetraj/im4}
%  \captionof{figure}{A human grasped a plate from the shelf and places it on the table. It can be observed how the gaze shifts towards the actual placing position even before the human reaches the table.}
%  \label{fig:place_plate}
%\end{center}}]

%%%%%%%%%%%%%%%%%%%%%%%%%%%%%%%%%%%%%%%%%%%%%%%%%%%%%%%%%%%%%%%%%%%%%%%%%%%%%%%%
\begin{abstract}
\input{abstract}
\end{abstract}

% Keywords appear just beneath the abstract. Use only for final RAL version. 

%%%%%%%%%%%%%%%%%%%%%%%%%%%%%%%%%%%%%%%%%%%%%%%%%%%%%%%%%%%%%%%%%%%%%%%%%%%%%%%%

\section{INTRODUCTION}
\input{introduction}

%\section{RELATED WORK}
%\input{rel-work}

\section{SETUP}
\input{setup}

\section{DATASET}
\input{dataset}

\section{APPLICATIONS}
\input{applications}

\section{SUMMARY AND FUTURE WORK}
\input{summary}

\section*{ACKNOWLEDGMENT}
The authors thank the International Max Planck Research School
for Intelligent Systems (IMPRS-IS) for supporting Philipp Kratzer.
This work was conducted while Simon Bihlmaier, Niteesh Midlagajni and Rohit Prakash
were performing their Bachelors and Masters dissertations in the Humans to Robots Motions Research group.

%%%%%%%%%%%%%%%%%%%%%%%%%%%%%%%%%%%%%%%%%%%%%%%%%%%%%%%%%%%%%%%%%%%%%%%%%%%%%%%%
\balance

\bibliographystyle{IEEEtran}
\bibliography{bibliography}

\end{document}

%% file: abstract.tex
As robots become more present in open human environments,
it will become crucial for robotic systems to understand and predict human motion.
Such capabilities depend heavily on the quality and availability of motion capture data.
However, existing datasets of full-body motion rarely include 
1) long sequences of manipulation tasks, 2) the 3D model of the workspace geometry,
and 3) eye-gaze, which are all important when a
robot needs to predict the movements of humans in close proximity.
Hence, in this paper, we present a novel dataset of full-body
motion for everyday manipulation tasks,
which includes the above.
The motion data was captured using a traditional
motion capture system based on reflective markers.
We additionally captured eye-gaze using a wearable pupil-tracking device. 
As we show in experiments, the dataset can be used for the design and evaluation
of full-body motion prediction algorithms.
Furthermore, our experiments show eye-gaze as a powerful predictor of human intent.
The dataset includes 180 min of motion capture data with
1627 pick and place actions being performed.
It is available at 
\href{https://humans-to-robots-motion.github.io/mogaze/}{\textit{MoGaze Dataset}}\footnote{\scriptsize{\url{https://humans-to-robots-motion.github.io/mogaze/}}}
and is planned to be extended to collaborative tasks with two humans in the near future.

%%% Local Variables:
%%% mode: latex
%%% TeX-master: "root"
%%% End:

%% file: introduction.tex
%Motivate
\IEEEPARstart{A}{s} robots become more capable,
space sharing robots will become a reality.
The involved close proximity to humans will lead to new challenges concerning
safety and acceptability of robots behaviors.
Especially robots that physically share space with humans
will need to understand and learn by interacting and observing humans.
Hence, intent and full-body motion prediction are becoming an increasingly
important topic of robotics research~\cite{rudenko2019human}.

This new trend raises the need for sophisticated datasets
that can be used to evaluate and design prediction algorithms.
In recent years with the development of deep-learning frameworks,
human-motion in crowded environment datasets such
as the recent Stanford Drone Dataset have become available~\cite{robicquet2016learning}. \revision{A recent dataset for robot autonomous navigation is JRDB~\cite{martin2019jrdb}. It includes data collected from a mobile data collection platform of indoor environments and pedestrian areas.}
These datasets typically include videos and tracking of pedestrians
in open-ended environments.
Recently, motivated by autonomous driving applications,
datasets recorded from a moving car perspective
such as~\cite{dollar2009pedestrian, geiger2013vision}
have become very popular.
However, for mobile manipulation application, where full-body
movements are more important, no such dataset is yet available.

Nonetheless, forecasting human whole-body motion as well as inferring
3D human pose has received significant attention from the computer vision community,
hence there is a large amount of available full-body motion datasets.
A widely used dataset is the CMU Graphics Lab motion capture database,
which contains a large amount of motion data from different humans~\cite{mocapcmu}.
Another commonly used dataset is the Human3.6m dataset
recorded by  Ionescu et al.~\cite{ionescu2013human3}.
It is a large scale dataset with 11 individual actors that is
often used to evaluate short-term motion prediction algorithms.
\revision{A large collection of short motion trajectories is the KIT Whole-Body Human Motion Database~\cite{mandery2015kit}. It can be used with a unified representation of human motion and motions can be transferred to humanoid robots~\cite{mandery2016unifying}.  The database contains motions including objects, such as drinking, shaking, pouring, and throwing of objects.}

More recent datasets include Marcard et al. who captured human 3D poses
in the wild by using a single hand-held camera and a set of Inertial Measurement Units~\cite{vonMarcard2018}. 
Ghorbani et al. presented a motion and video dataset
that can be used for body shape estimation~\cite{ghorbani2020movi}.
Maurice et al. captured an industry-oriented dataset recorded both
with wearable inertial sensors~\cite{maurice2019human}.
In this recent dataset, the participants performed activities,
such as screwing and manipulating loads, which are often found
in collaborative robotics scenarios.

\begin{figure*}[!t]
  \centering
  \begin{subfigure}{.345\textwidth}
  \centering
  \includegraphics[height=4.5cm]{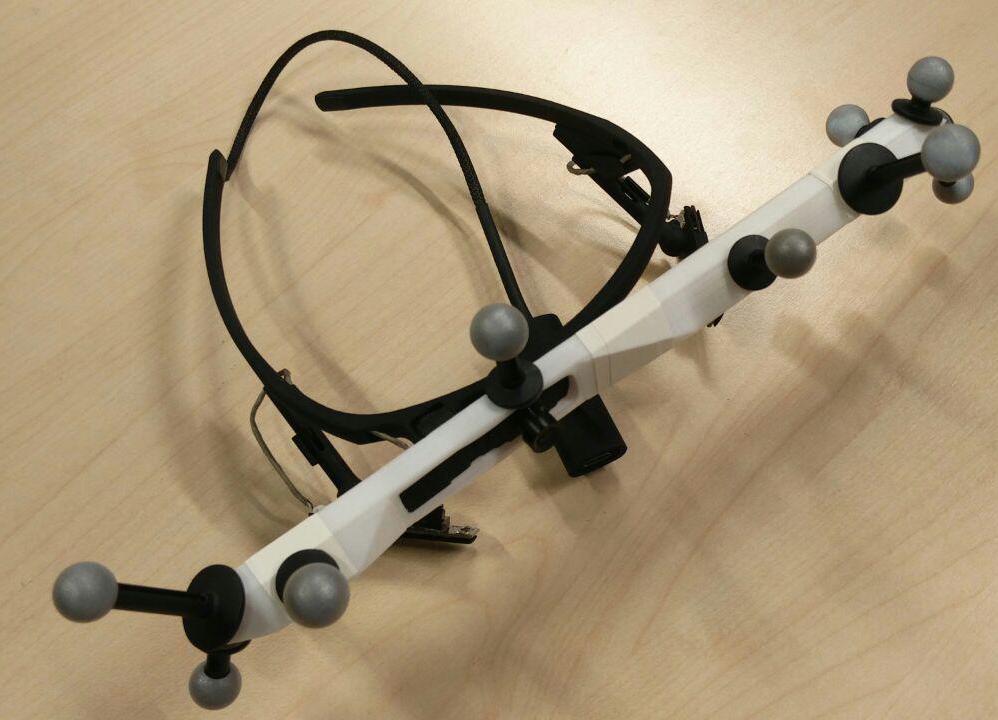}
  %\label{fig:attachment_real_foto}
  \end{subfigure}
  \begin{subfigure}{.295\textwidth}
    \centering
    \includegraphics[height=4.5cm]{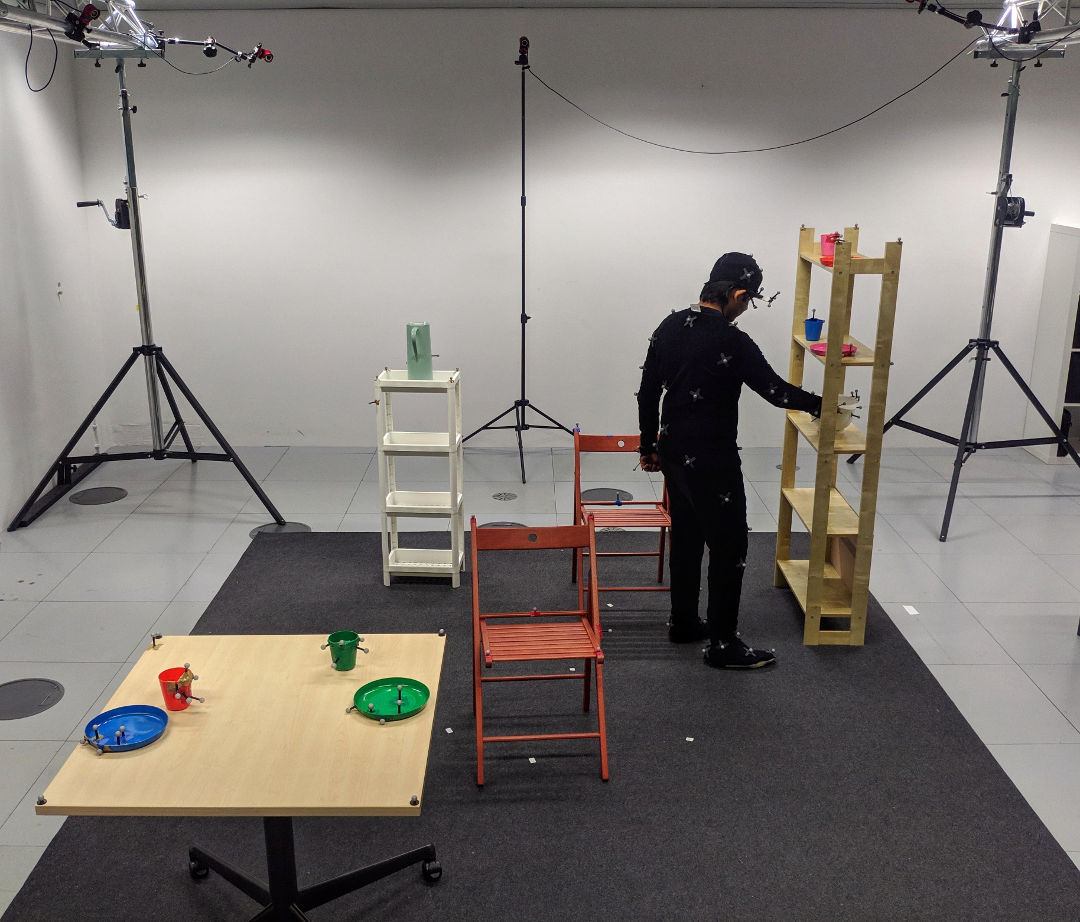}
  \end{subfigure}
  \begin{subfigure}{.32\textwidth}
    \centering
    \includegraphics[height=4.5cm]{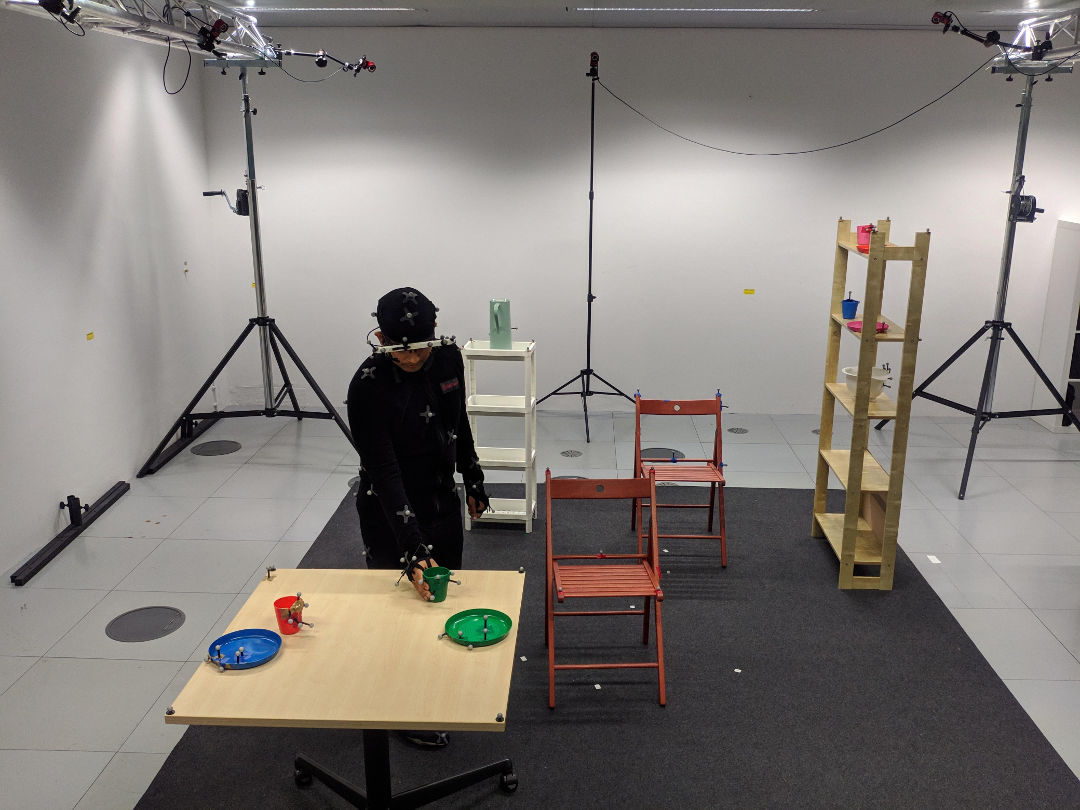}
  \end{subfigure}
  \caption{Eye tracking glasses with removable 3D printed attachment for motion capture markers (left) . The motion capture scene with a human participant and objects (right).}
  \label{fig:mocapscene}
  \vspace{-.5cm}
\end{figure*}

Interestingly,
Mahmood et al. presented a dataset that aggregates multiple datasets, including
some of the above into a common parameterization
called AMASS~\cite{mahmood2019amass}.
Motion capture datasets are usually captured using different
systems and marker sets. AMASS's  achieved a single parameterization
by converting marker data into realistic 3D human meshes
represented by a rigid body model based on 4D motion scans.
The original dataset contains 40h of marker data,
which makes it the largest publicly available full-body human motion database to date.
This is desirable for deep-learning approaches, as it limits
over-fitting and allows to generalize across humans.

\revision{Motion prediction is very important for robotics applications.
Many works tackle the crowd movement problem.
For example, Curtis et al. presented \textit{Menge}, a framework that can
be used to simulate pedestrian movement in crowds~\cite{curtis2016menge}.
Fan et al. proposed a navigation
framework ~\cite{fan2019getting}
that addresses robots getting frozen or lost in dense crowds,
such as in shopping malls by using motion prediction.}

However, in service robotics applications,
where a robot \revision{intimately} shares space with one or multiple moving humans,
the intent of the humans is often dictated by the task that they are performing.
For instance in a household, restaurant or hospital environment,
humans are often transporting or manipulating objects in their surroundings.
Hence, here we aim to capture long sequences of pick and place
movements, that we think are representative of the task that
a service robot would encounter, which may or may not be organized hierarchically.

For predicting and understanding movement in such cases,
the 3D geometry of the workspace is key. This is
also absent from currently available datasets. Having
access to the geometry of the workspace will allow
to model how humans perform motion planning in unstructured environments.

% Motivate gaze + RW gaze
Humans often start looking toward an object before they start to move~\cite{admoni2017social}.
This indicates that gaze features are a useful indicator for intent prediction and early detection of human activity.
In related work, for instance, Huang and Mutlu performed an experiment,
where participants had to order items from a robot.
Using data from eye tracking, their system uses early prediction to make the robot proactively support the human~\cite{huang2016anticipatory}.
In another experiment for action anticipation,
Duarte et al. investigated nonverbal cues and found that eye gaze provides
the key information that helps humans identify actions of other humans correctly~\cite{duarte2018action}.

As early prediction of human intent and full-body motion prediction can be combined~\cite{kratzer2020anticipating}, investigating both subjects together makes sense. However, to our knowledge there are no datasets that include both, human full-body motion capture data and gaze.

%\begin{figure}
%  \includegraphics[width=\columnwidth]{human_scene}
%  \caption{Human looking towards a plate on the table.}
%  \label{fig:human_scene}
%\end{figure}

% RW datasets

\begin{figure*}[!t]
  \centering
  \includegraphics[width=.83\textwidth]{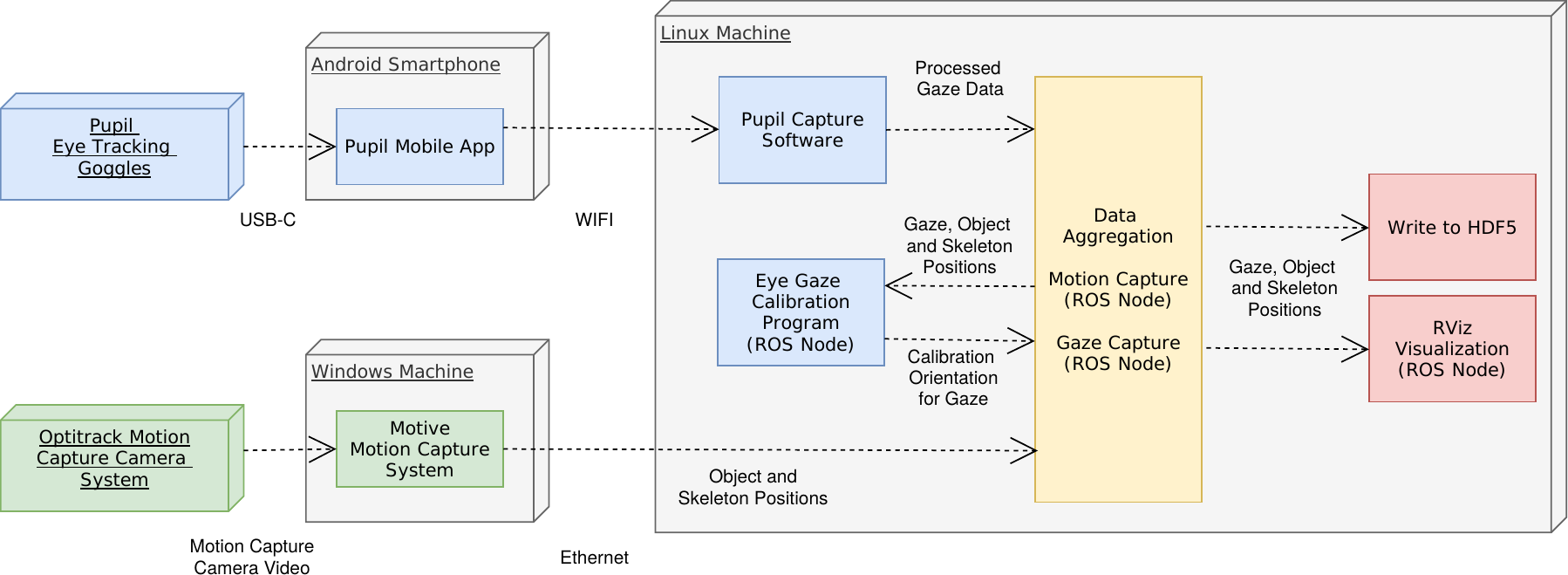}
  \caption{
  Dataflow of our full-body motion \& eye-gaze capture framework.
  Four types of modules (colored) are distributed over three machines (grey):
  full-body motion capture (green), 
  eye-gaze capture (blue), data aggregation (yellow) and data visualization and saving (red). }
  \label{fig:capturesetup}
  \vspace{-.4cm}
\end{figure*}

To summarize the main issues concerning the applicability of the available
dataset to mobile manipulation \& service robotics are that they:
\begin{itemize}
\item do not consider manipulation tasks
\item do not include 3D scene/workspace geometrical context 
\item do not include gaze.
\end{itemize}

% paper overview

In this work, we present a dataset of human acting
in a table-setup and cleaning tasks, which addresses these shortcomings.
The presented dataset is the first to our knowledge that includes both:
full-body motion data and gaze.
Figure~\ref{fig:place_plate} depicts a sample trajectory and the scene we consider.
The dataset includes multiple objects, such as tables, shelves, plates, and cups, as well as full-body motion data of 7 different actors and eye-gaze data for 6 of the actors.

%%% Local Variables:
%%% mode: latex
%%% TeX-master: "root"
%%% End:

%% file: setup.tex
We used an OptiTrack\footnote{\scriptsize{\url{https://www.optitrack.com/}}}
motion capture system with 12 cameras.
The cameras track an area of approximately 3 by 4 meters
as can be seen in Figure~\ref{fig:mocapscene}.

\paragraph{Human movement}
The human participants wore a motion capture suit with
50 reflecting markers placed on the body including arms and legs.
The marker set follows the convention for full-body tracking in Motive
\footnote{\scriptsize{\url{https://www.optitrack.com/products/motive/}}} configuration.
This allows good tracking of human full-body motion.

\paragraph{Workspace geometry}
Four to six markers are 
placed on the objects in the scene so that the position and orientation
of the rigid-body corresponding to each object,
including plates, cups, jug, bowl, shelve, tables, and chairs, can be estimated.
All dimensions of the objects have been measured and
the geometries were modeled as triangular meshes using CAD software.
The data is captured at a framerate of 120Hz.

\paragraph{Eye-gaze}
For eye tracking we use a headset from PupilLabs
\cite{Kassner:2014:POS:2638728.2641695},
which uses two eye cameras pointing inwards to estimate the pupils' motion
and a world camera to calibrate the estimated gaze direction
with respect to the headset.
A 3D printed removable attachment was added
to track the headset pose in the motion capture volume.
This allows to accurately track the gaze direction in global coordinates.
The headset with the attachment clipped onto the world camera can be seen in Figure~\ref{fig:mocapscene}.

An overview of the capture system can be seen in Figure~\ref{fig:capturesetup}.
The eye tracking headset is connected to an Android smartphone via USB-C.
This allows the human to move freely in the scene without being
hindered by a tether.
The smartphone runs the Pupil Mobile Android app, which uses  Wi-Fi to send the video of the eyes of the human and the world camera to the Pupil Capture Software running on a Linux workstation.
The motion capture system is connected to a Windows workstation which runs the Motive motion capture software. The Windows machine streams the motion data to the Linux workstation.
On the Linux machine, programs for recording the gaze data and the motion capture data are running. The live data is visualized and simultaneously written to HDF5 files.

The eye-tracker's internal clock is synchronized
with the clock on the Linux workstation by using scripts provided by PupilLabs.
The motion capture data is also stamped
with the clock of the Linux workstation. \revision{The mean delay between the Windows machine and the Linux machine is 0.00055s, which is negligible.}
In a post-processing phase, we use the stamps to synchronize
the gaze and motion capture frames. The motion data is captured with a frequency of 120Hz and the gaze with a frequency of 200Hz. When multiple gaze data frames correspond to one motion capture frame,
we use the frame with a higher reported confidence value from the Pupil software.

\subsection{Calibration of Motion Capture and Eye-Tracker}

For calibrating the motion capture system and the eye-tracker,
we use the calibration routines provided by the manufacturer.
The motion capture system is calibrated by capturing markers 
mounted on a rigid object moved inside the volume.
The eye-tracker is calibrated by letting the participant look at visual markers
at different locations, which are tracked by the headset world camera.
This calibrates the pupils position to the gaze point on the image plane of the headset's camera.

In order to obtain the gaze-vector in global coordinates,
we additionally need the pose of the headset's world camera in
the global coordinates of the motion capture system,
which is why we add markers to the headset (see Figure~\ref{fig:mocapscene}).
However, errors can still exist between the pose of the rigid-body 
estimated by the motion capture system and the true pose of the headset's world camera.
Hence to circumvent this, we directly calibrate the
gaze in global coordinates by letting the participant look at a predefined
motion capture marker several times from different angles.
We then perform a least squares fit to calculate the
pose of the headset's world camera.
% -----------------------------------------------------------------------------------------------------

%%% Local Variables:
%%% mode: latex
%%% TeX-master: "root"
%%% End:

%% file: dataset.tex
In total, 180 minutes of data were recorded. Capturing was done with seven participants. Six participants were male, one was female. All the participants were students.  For one participant the eye-tracker device did not work due to his glasses, so that no gaze data is provided for this participant.

\subsection{Procedure}

Each of the participants received a short introduction on how the capture setup works, what the tasks are and what they imply. For a description of the tasks, see Subsection~\ref{ssec:tasks}.
The participants were advised to only use one hand for performing the task and only move one object at a time.

The motion capture suit was put on and the reflective markers were adjusted.  The eye-tracker was connected to the smartphone and calibrated.
We performed a short test recording to get the participants used to the capture setup and the tasks.
Then, we started the recording session. In case of recording issues the session was interrupted and resumed.

\subsection{Scene and Objects}
\begin{figure}
  \newcommand{\ltscale}{.15}
  \begin{subfigure}{\ltscale\textwidth}
    \includegraphics[width=\linewidth]{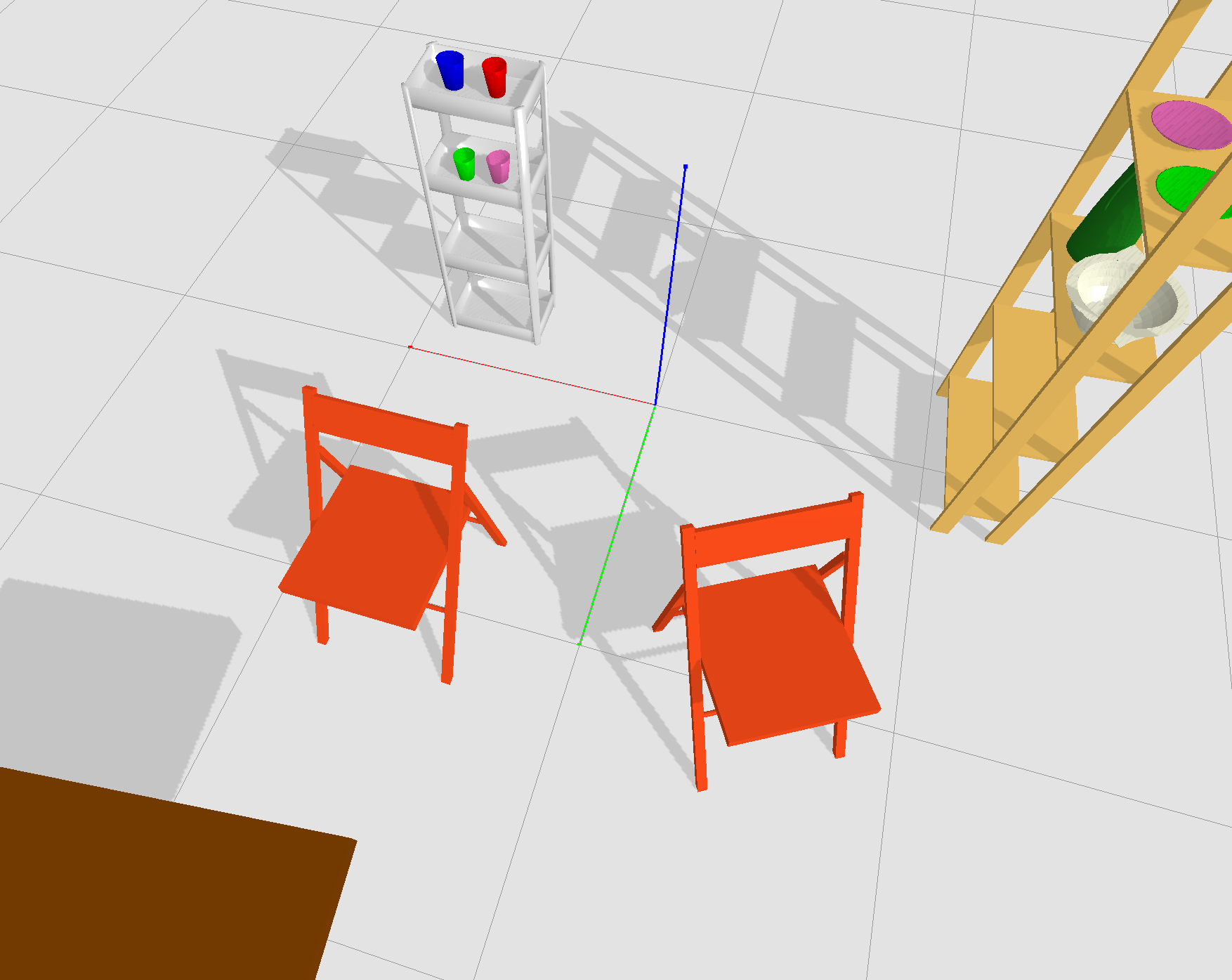}
    \caption{Configuration 1}
    \label{fig:chairconfig1}
  \end{subfigure}
  \hfill
  \begin{subfigure}{\ltscale\textwidth}
    \includegraphics[width=\linewidth]{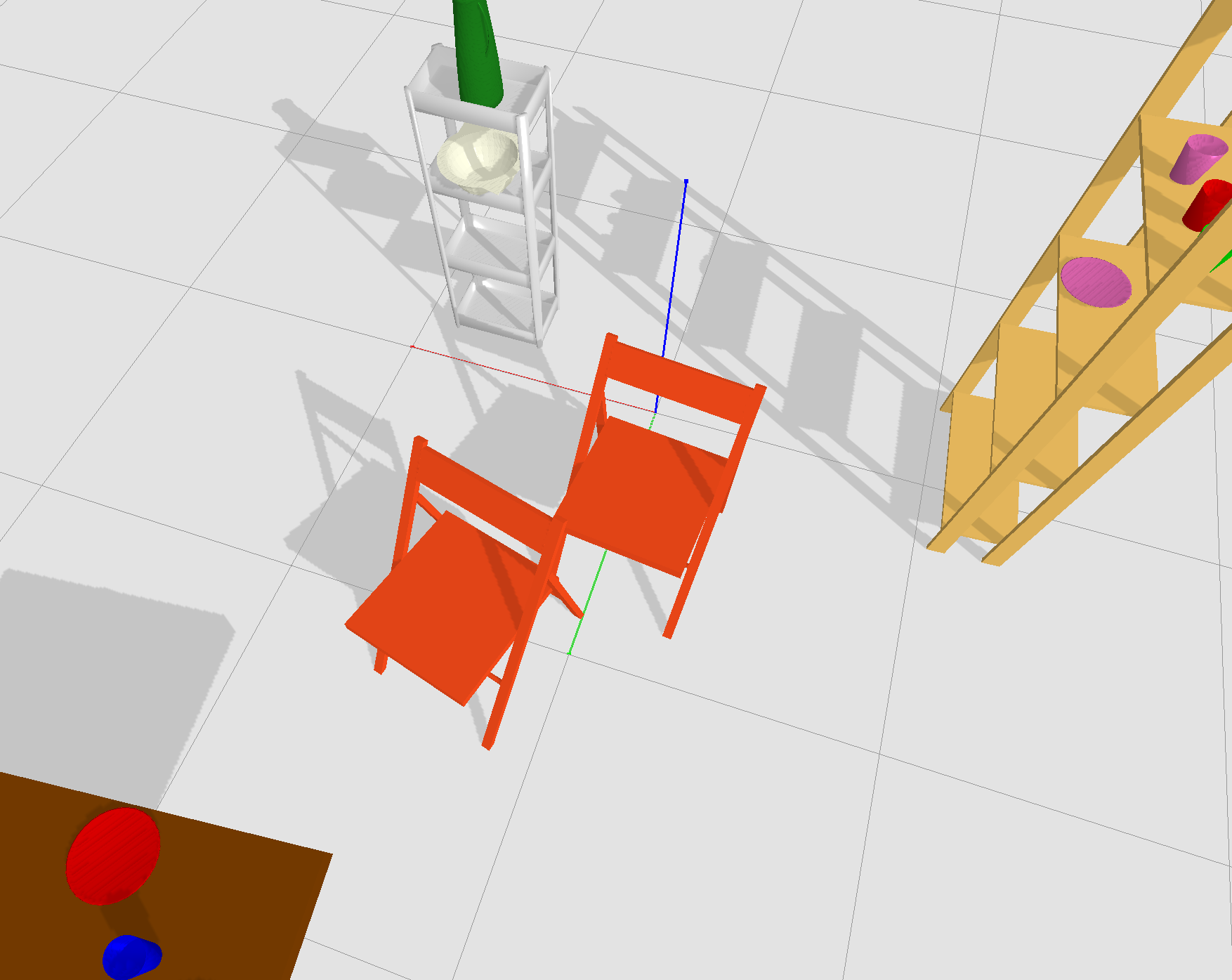}
    \caption{Configuration 2}
    \label{fig:chairconfig2}
  \end{subfigure}
  \hfill
  \begin{subfigure}{\ltscale\textwidth}
    \includegraphics[width=\linewidth]{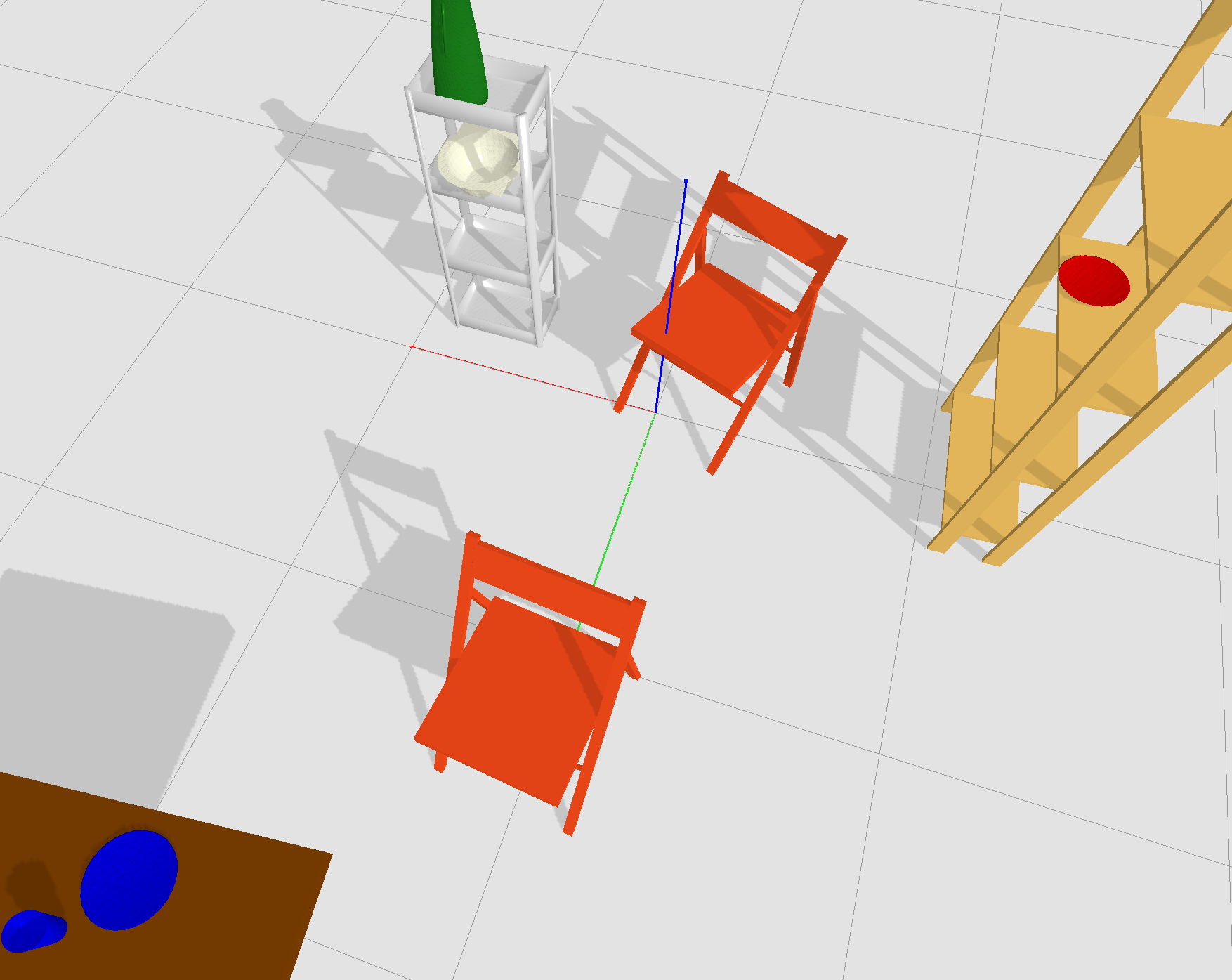}
    \caption{Configuration 3}
    \label{fig:chairconfig3}
  \end{subfigure}
  \caption{Different configurations of the chairs}
  \label{fig:chairconfigs}
\end{figure}

The following objects were available in the scene:
4 plates and 4 cups (colored red, green, blue and pink), 1 jug, 1 bowl, 2 chairs, a big shelf, a small shelf and a table.
The shelves and the table were kept at a fixed position for the whole recording. The plates, cups, jug, and bowl were moved around by the participants during the tasks.

The positions of the chairs changed three times throughout the recording, in order to prevent participants
from always taking the same paths and thus collect more variations of motion.
The different configurations can be seen in Figure~\ref{fig:chairconfigs}. \revision{The configurations are designed in a way that they lead to different motion behaviors for the human. While the first and the last configuration allow the human to walk in between the chairs, in the second configuration the chairs are placed close to each other so that the human needs to walk a longer route around them. These three diverse configurations are used for every participant. We chose three configurations because this is a good trade-off between getting more variation into the walking behavior and having sufficient data per configuration.}

\subsection{Tasks}
\label{ssec:tasks}
\begin{table}
	\centering
	\begin{tabular}{l l}
		\textbf{Probability} & \textbf{Task} \\
		0.08\revision{1}  &  Set the table for 1 person \\	
		0.08\revision{1}  &  Set the table for 2 persons \\
		0.08\revision{1}  &  Set the table for 3 persons \\
		0.08\revision{1}  &  Set the table for 4 persons \\
		0.03\revision{2}  &  Clear table \\
		0.16\revision{1}  &  Put the jug and the bowl on small shelf \\
		0.16\revision{1}  &  Put all cups on small shelf \\
		0.03\revision{2}  &  Put blue and pink objects on big shelf \\
		0.03\revision{2}  &  Put blue and red objects on big shelf \\
		0.03\revision{2}  &  Put blue and green objects on big shelf \\
		0.03\revision{2}  &  Put pink and red objects on big shelf \\
		0.03\revision{2}  &  Put pink and green objects on big shelf \\
		0.03\revision{2}   &  Put red and green objects on big shelf \\
		0.03\revision{2}   &  Put all cups on big shelf \\
		0.03\revision{2}   &  Put bowl and jug on big shelf \\
		0.03\revision{2}   &  Put all cups on big shelf \\
		0.03\revision{2}   &  Put all plates on big shelf \\
	\end{tabular}
	\caption[Instruction table]{Table containing the instructions that were used, paired with the probability of the task being assigned to the participant}
	\label{tab:Instructions}
	\vspace{-.5cm}
\end{table}  
During the recording session, the participants had to perform simple manipulation tasks. A full list of the tasks can be seen in Table~\ref{tab:Instructions}. The tasks were to either set up the table for a specific amount of people or to move a certain group of objects to a specific shelf.

For setting up the table, the corresponding number of plates and cups and either the jug or the bowl had to be placed on the table.  In order to balance the locations, to which objects are moved to, an occurrence probability was attached to the tasks.
\revision{The tasks are balanced so that a task for moving
objects to the table, moving objects to the big shelf, or moving objects to the small shelf
is selected with equal probability.
This ensures that the objects are equally distributed in the area.}

An instruction program randomly selects a task according to the corresponding
probabilities and shows it to the conductor of the experiment.
The experiment conductor would then decide if the task is possible in the current state of the scene.  After accepting, the task is read out to the participant using a text-to-speech system.

The instruction program stores the tasks during data capturing and stamps it, which makes it possible to know which task the participant was performing at every time step in the recording.

\subsection{Labeling of the data}
For some applications, it is important to index the data as picking or placing movement. 
We automatically label the data by calculating when an object moves and stops moving.
This was done by comparing the positions of the object to previous timesteps
and consider it as moving when a threshold is passed.
The labeling is validated and corrected by a human expert.
%When an object starts to move a pick action is performed by the human,
%when the movement stops a place action is performed.

\begin{figure}
  \centering
  \includegraphics[width=\columnwidth]{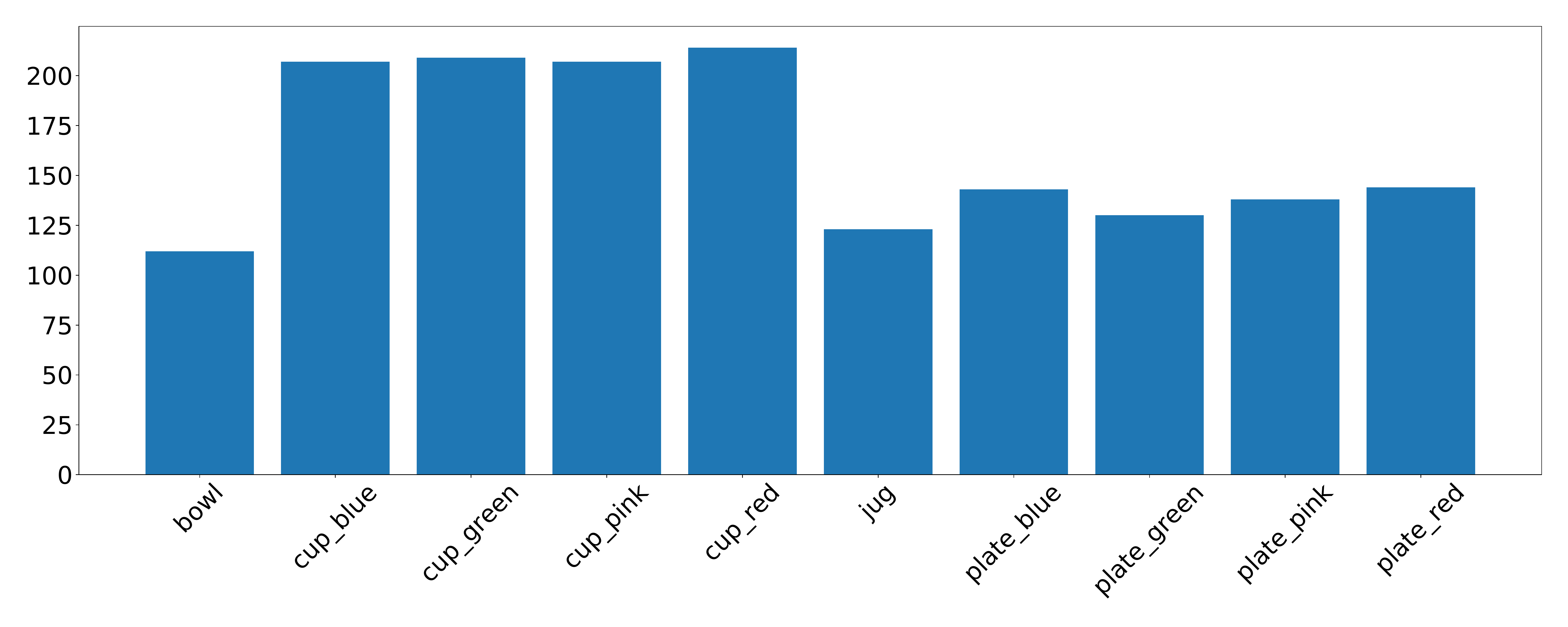}
  \caption{Number of pick and place actions per object.}
  \label{fig:counts}
    \vspace{-.5cm}
\end{figure}

Figure~\ref{fig:counts} shows the number of pick and place actions per object. In total 1627 pick and place actions have been performed. The cups are the objects, which are most often moved, which is expected as they occur in most of the tasks. Objects of the same type, such as cups or plates, have a similar number of actions.

 \begin{figure}
   \centering
   \includegraphics[width=\columnwidth]{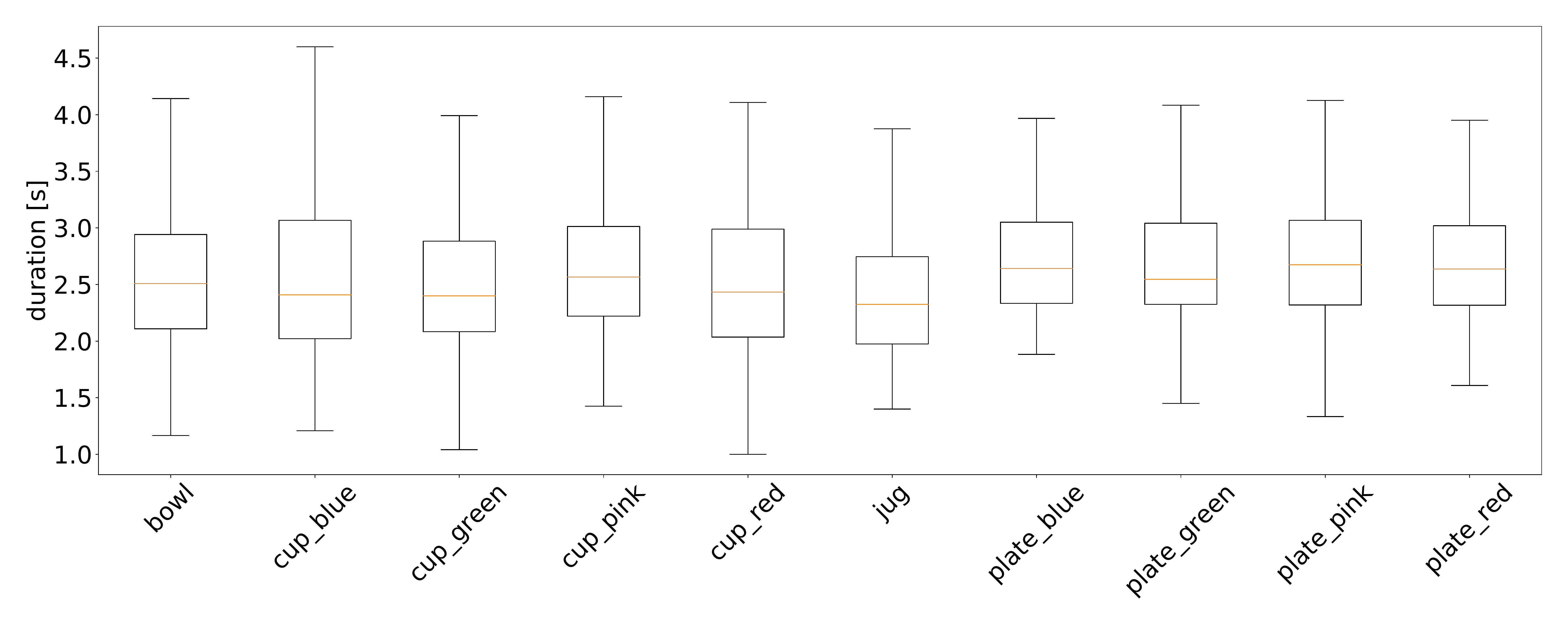}
   \caption{Placing durations, from object is grasped till object is placed, per object. }
   \label{fig:durations}
     \vspace{-.5cm}
 \end{figure}

 Figure~\ref{fig:durations} depicts the duration of the grasping action, which is the duration from when an object is picked till it is placed.  It can be seen that the timespan of an action is usually between two and three seconds. This holds for all the objects in the dataset.

\subsection{Dataset Structure and Contents}
%\begin{figure}
%  \centering
%  \includegraphics[width=.7\columnwidth]{bounding_boxes.png}
%  \caption{Bounding boxes of the movable objects}
%  \label{fig:bounding_boxes}
%\end{figure}
%
% \begin{figure}
%   \centering
%   \includegraphics[width=.9\columnwidth]{sdf.png}
%   \caption{Occupancy and signed distance field features
%   of the chairs, table and shelves projected on the plane. }
%   \label{fig:sdf}
% \end{figure}

%  \begin{figure*}[h!]
%   %\centering
%   \newcommand{\ltscale}{.45}
%    \begin{subfigure}{\ltscale\textwidth}
%   \includegraphics[width=.56\linewidth]{sdf_gimp.png}
%   \includegraphics[width=.35\linewidth]{bounding_boxes2.png}
%   \caption{Occupancy map and signed distance field features maps
%   of the chairs, table and shelves projected on the plane (left), bounding box of object on shelf (right). }
% \label{fig:sdf}
% \end{subfigure}
% \hfill
%   \begin{subfigure}{\ltscale\textwidth}
%    \centering
%   \includegraphics[width=.5\linewidth]{gaze_intersection_point.png}
%   \caption{\revision{Gaze intersection point feature. The point of intersection is on the green cup on the table.}}
%   \label{fig:gint}
%     %\vspace{-.7cm}
% \end{subfigure}
% \caption{Features maps}
%   \vspace{-.2cm}
% \end{figure*}

  \begin{figure}[t]
    \centering
   \includegraphics[width=.56\linewidth]{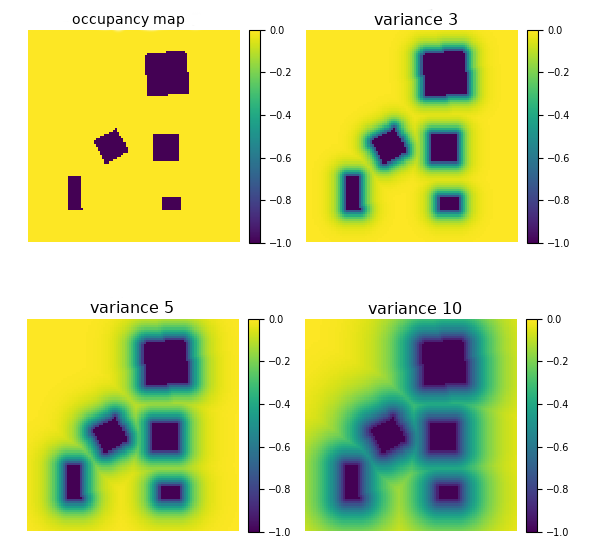}
   \includegraphics[width=.34\linewidth]{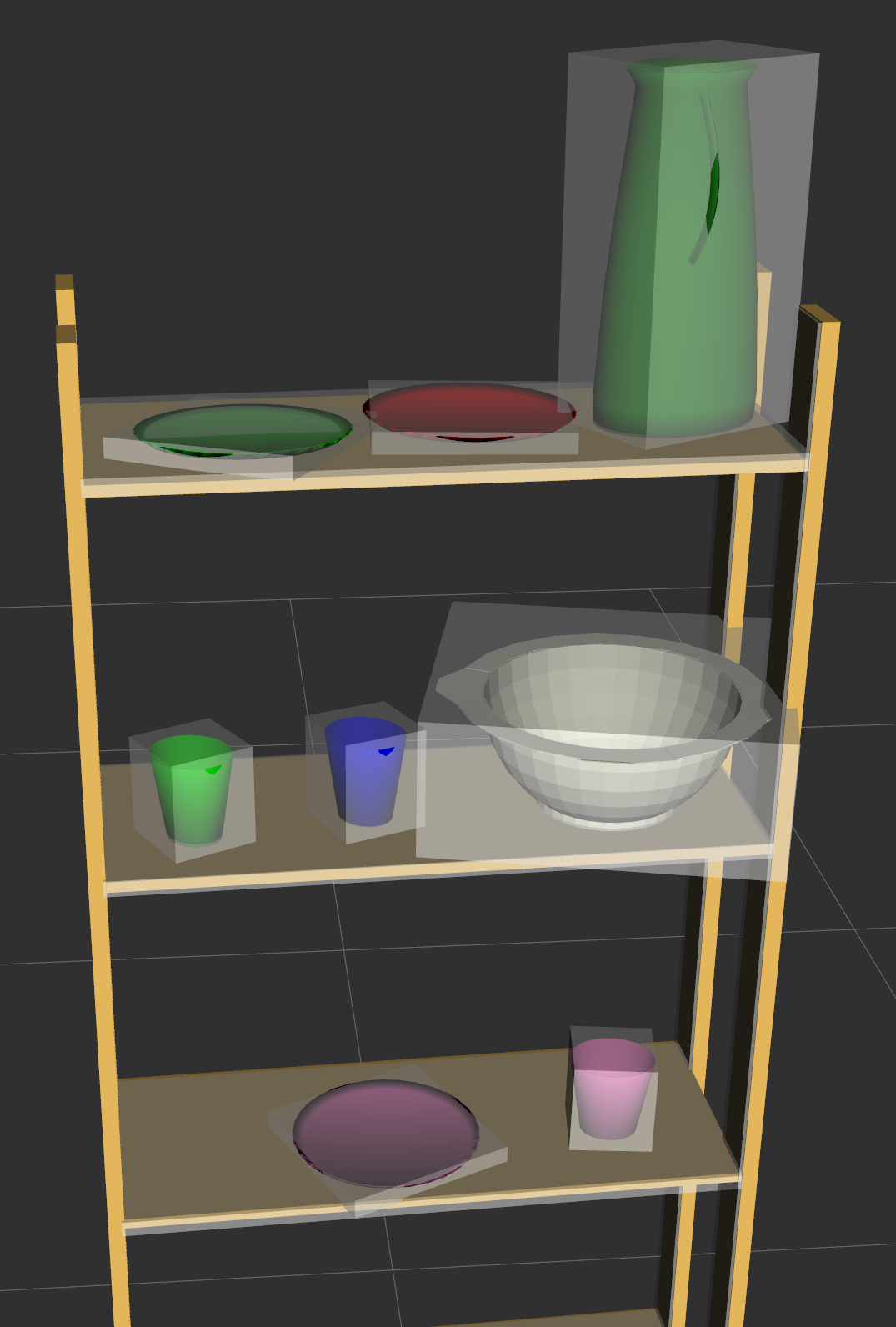}
   \caption{Occupancy map and signed distance field features maps
   of the chairs, table and shelves projected on the plane (left), bounding box of object on shelf (right). }
   \label{fig:sdf}
%     \vspace{-.7cm}
 \end{figure}

  \begin{figure}[t]
    \centering
   \includegraphics[width=.55\columnwidth]{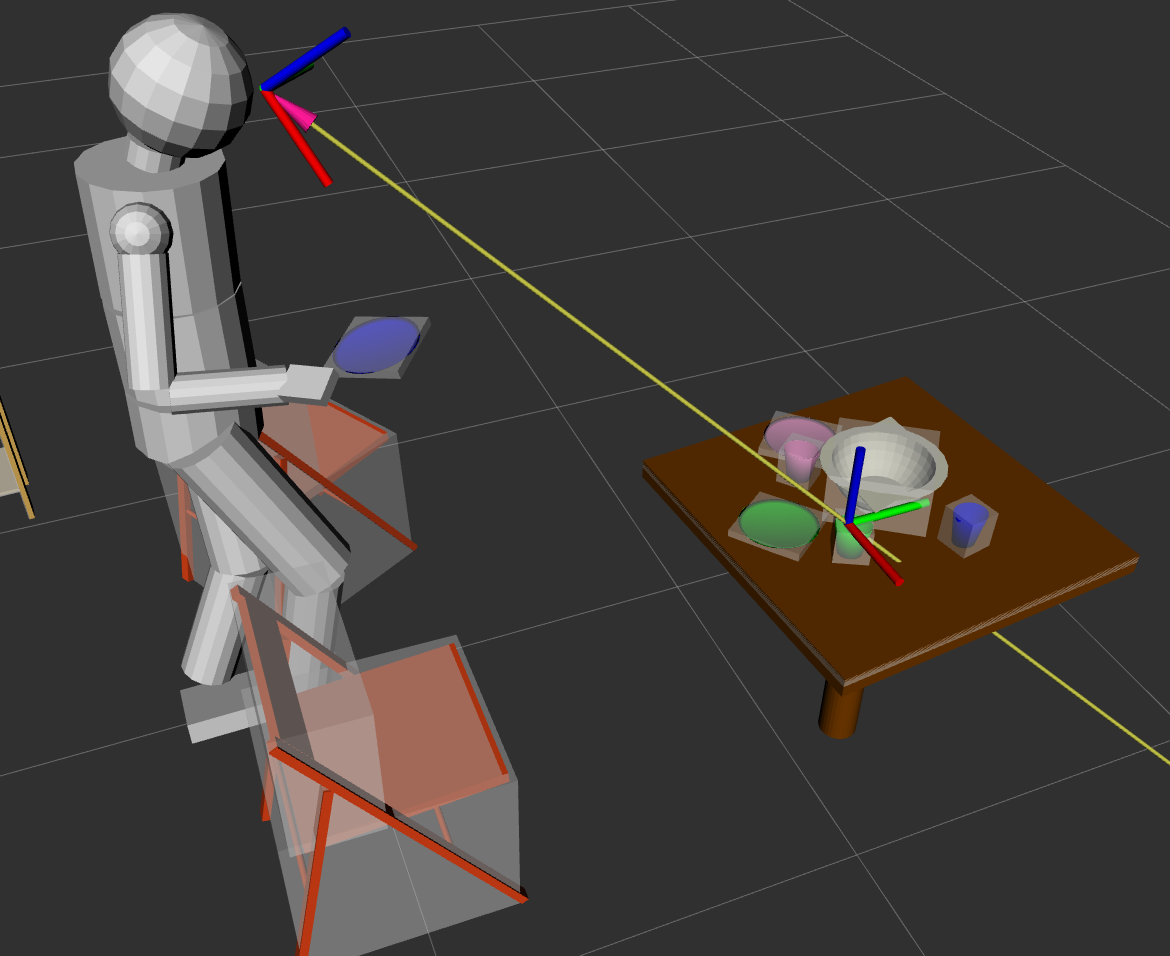}
   \caption{\revision{Gaze intersection point feature. The point of intersection is on the green cup on the table.}}
   \label{fig:gint}
     \vspace{-.5cm}
 \end{figure}

 \revision{
   \paragraph{File Formats}   
   The recorded data of the human, the objects, and the gaze are stored as \textit{hdf5} files. The \textit{hdf5} file format is designed for storing large amounts of data and for fast input/output processing. The instructions are stored in a \textit{csv} file because the files are smaller and include human readable text. A scene description, containing per object properties, such as bounding boxes or names of mesh files, is stored in the \textit{xml} file format. A \textit{urdf} model for the human skeleton is provided.} Some of the data is post-processed to map the gaze frames to the motion frames or generate segmentations of the data.
 
 \paragraph{Available Contents}
 
The following information is available for all recordings:
\begin{itemize}
	\item Positions and orientations of the objects, including the eye-tracker
	\item Base position and joint angles of the human skeleton
	\item Gaze data from the eye-tracker and calibration rotation
	\item Tasks and the timestamps at which they were posed to the participant
        \item Segmentation data into pick and place
        \item Raw motion capture marker data
\end{itemize}	
Additionally, we provide mesh files of the objects in the scene for
visualization.

\revision{\paragraph{Data Structure}
  The human data \textit{hdf5} file has a dataset containing the joint state vector of the human for every frame. In the attributes it has a \textit{description} array of strings with the names of the corresponding joint name of the human model. Additionally, the attributes contain a list of fixed joints, which are only used to scale the \textit{urdf}  to different humans, with a corresponding value.
  
  The object data \textit{hdf5} file contains a single database for every object containing a seven dimensional vector with position and quaternion orientation for every frame.
  The gaze file contains a vector of gaze data from the pupil eye-tracker per frame. The data includes a 2D gaze position in the image plane, a 3D gaze point, the eye center coordinates for both eyes, the gaze normals for both eyes, and a confidence value.
  
  The instruction \textit{csv} file has three columns for the frame when the instruction was read, the id of the task, and the description of the task.
  
  The segmentation file contains a database of tuples with the start frame of the segment, the end frame of the segment, and the segment name.
}

We provide a library to visualize and playback all the data based on PyBullet~\cite{coumans2019}.

\subsection{Features}
To use the data in applications, several features \revision{regarding objects, the human, and gaze} can be computed with the dataset.

Figure~\ref{fig:sdf} depicts examples of \revision{object} features. On the left, a 2D occupancy of the environment and \revision{several exponentiated signed distance field feature maps with different variances are shown. We will introduce an application that uses them in Subsection~\ref{ssec:HMP}. The right side shows a box representation of objects. It can be used as collision shapes or for 3D signed distance field computation.

  The gaze features include the distance of objects to the gaze-ray and gaze intersection points with objects as seen in Figure~\ref{fig:gint}. This can also be used to calculate which object the human currently looks at. An application that uses gaze features is introduced in Subsection~\ref{ssec:gaze}.

For human features, the data is available in a joint angle representation. Positional features, such as positions and rotations of joints or links, can be computed using forward kinematics. This also can be used to compute human-object features, such as the distance of the hand to an object.}

\subsection{Limitations}
The cameras were adjusted to cover the full scene. However, due to the various number of objects and shelves, occlusions can rarely occur. This can lead to small jumps of objects.

The eye-tracker needs to be calibrated for a specific distance to the eyes. As the human is moving in the scene, the distance between eyes and objects he or she is looking at varies. We calibrated the eye-tracker for a distance of approximately one meter because we found that this is a good compromise for our tasks.

%%% Local Variables:
%%% mode: latex
%%% TeX-master: "root"
%%% End:

%% file: applications.tex
%\begin{figure}
%  \centering
%  \includegraphics[width=\columnwidth]{prediction1.png}
%  \caption{Prediction of a human reaching towards a bowl on the shelf.}
%  \label{fig:prediction}
%\end{figure}

\begin{figure*}
  \newcommand{\ltscale}{.33}
  \begin{subfigure}{\ltscale\textwidth}
    \centering
    \includegraphics[width=\linewidth]{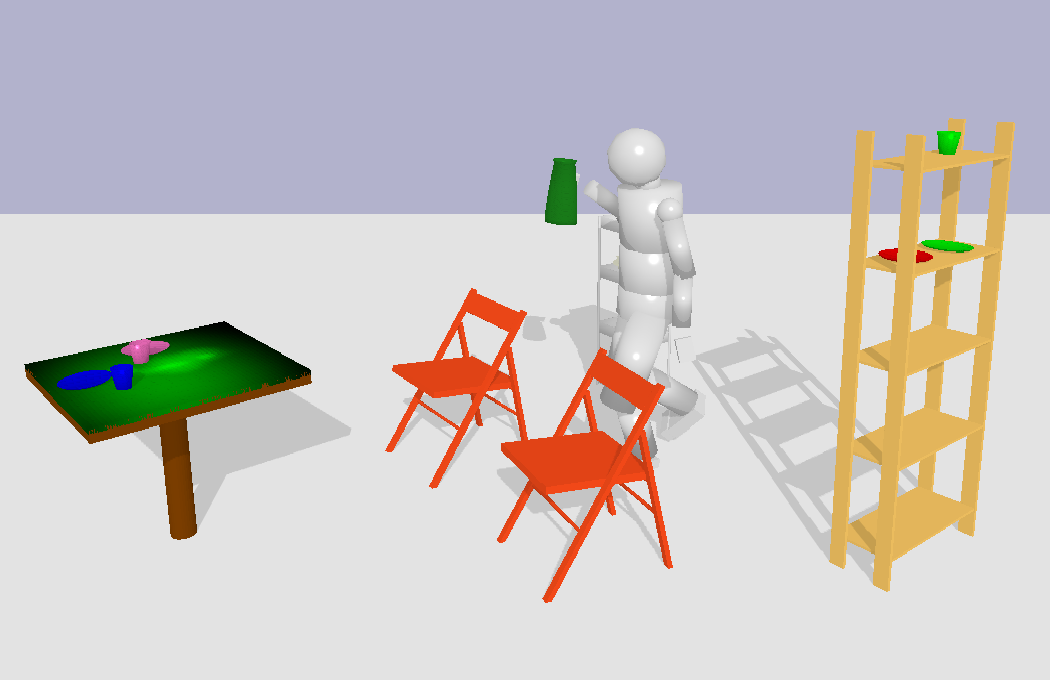}
    \caption{Initial state with affordance heatmap}
    \label{sfig:heatmap}
  \end{subfigure}
  \begin{subfigure}{\ltscale\textwidth}
    \centering
    \includegraphics[width=\linewidth]{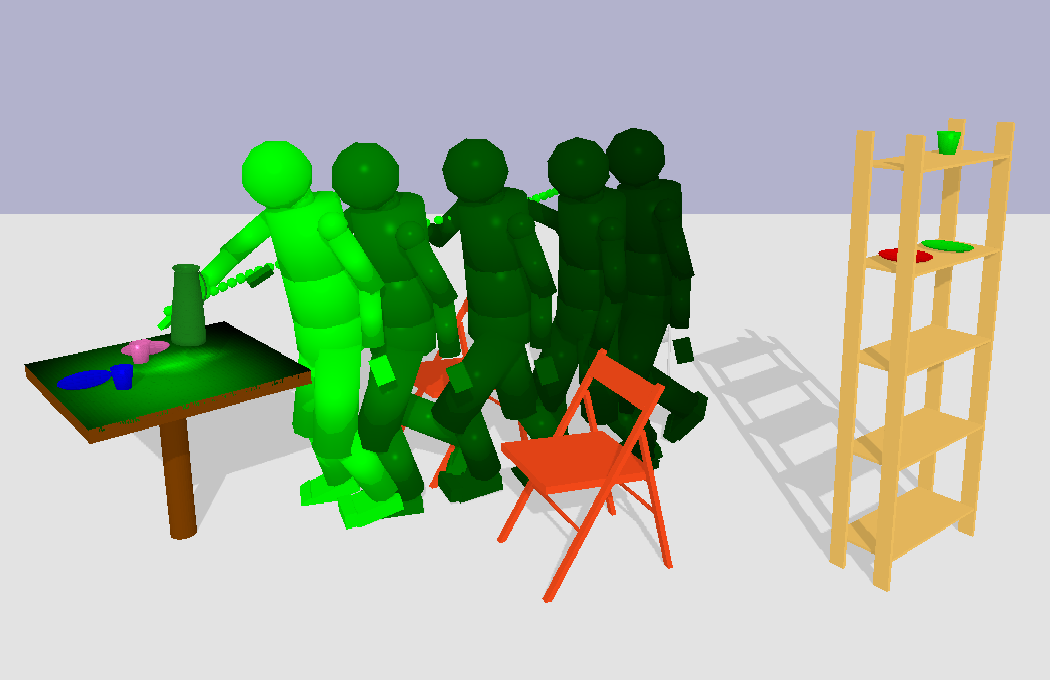}
    \caption{Full-body prediction using affordance}
    \label{sfig:fbaffordance}
  \end{subfigure}
  \begin{subfigure}{\ltscale\textwidth}
    \centering
    \includegraphics[width=\linewidth]{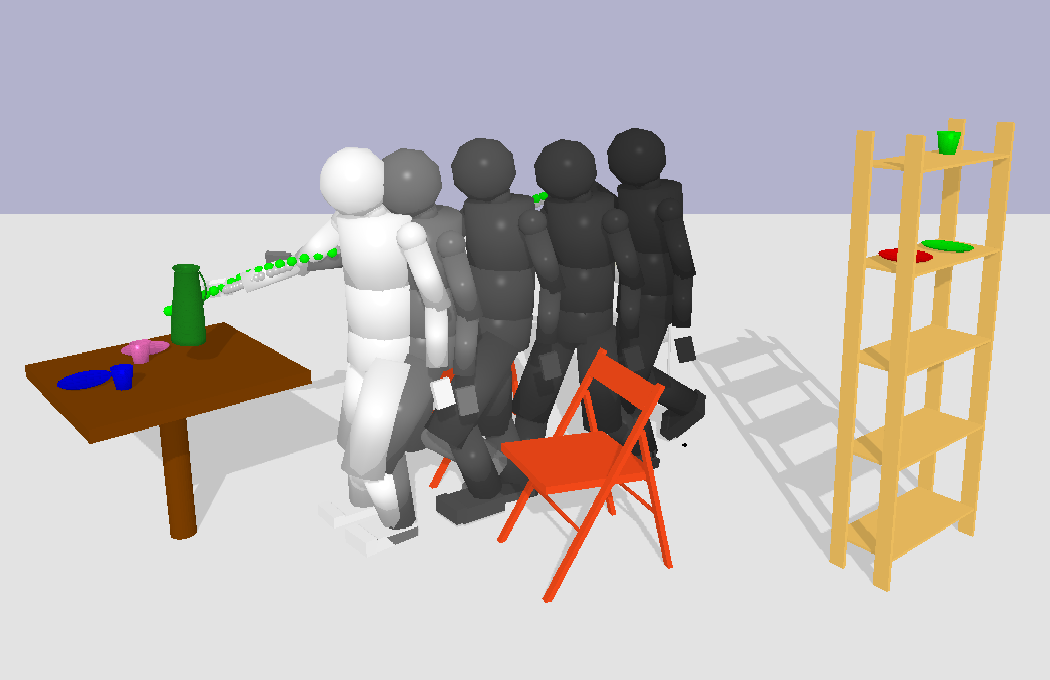}
    \caption{Ground truth trajectory}
    \label{sfig:fbgt}
  \end{subfigure}
  \caption{Prediction of placing a jug. A placement affordance is predicted as a probability density function on the table (a), depicted in green a full-body motion is optimized (b), which is compared to ground truth motion (c).}
  \label{fig:place_jug}
  \vspace{-.5cm}
\end{figure*}

%\begin{figure}
%   \centering
%   \includegraphics[width=.9\columnwidth]{place_auto.png}
%   \caption{Placeability network architecture}
%   \label{fig:place_auto}
% \end{figure}
% 
 \begin{table}
  \begin{tabular}{l|l|l}
    S.No. & States & Meaning \\
    \hline
1. & [2, 2, 2, 2, 1, 1, 1, 1, 1, 1, 2] & Initial state              \\
2. & [0, 2, 2, 2, 1, 1, 1, 1, 1, 1, 0] & cup from small shelf to table\\
3. &[0, 2, 2, 2, 0, 1, 1, 1, 1, 1, 0] &  plate from big shelf to table\\
4. & [0, 2, 2, 2, 1, 1, 0, 1, 1, 0, 0] & jug from big shelf to table
  \end{tabular}
  \caption{Example of high-level state trajectory, \revision{with state number, a high-level state representation and meaning. The high-level state representation describes where an object (first ten elements) or the human (last element) is located (0=table, 1=big shelf, 2=small shelf).}}
  \label{tab:hltrajectory}
\end{table}

 \begin{figure}[t]
   \centering
   \includegraphics[width=\columnwidth]{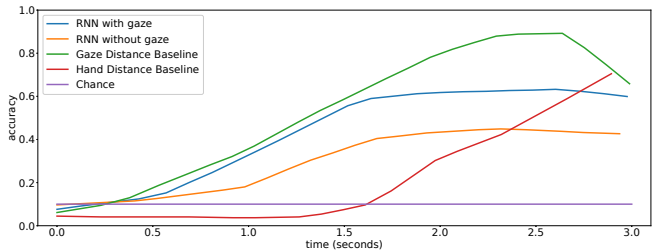}
   \caption{Prediction accuracy of which object is being picked. A simple gaze-distance based method outperforms other models. }
   \label{fig:pickpred}
   \vspace{-.4cm}
 \end{figure}

In the following, we briefly describe applications performed with the dataset or a subset of the dataset. The experiments demonstrate the benefit of the dataset for the research field of human motion prediction.

\subsection{Context-Aware Full-Body Motion Prediction}
In~\cite{kratzer2019prediction} we use a subset of the dataset to train a recurrent neural network based short-term prediction model for full-body human motion. The aim is to find a trajectory of a human motion $h_{t+1:T}$, given a trajectory $h_{0:t}$ of previously observed states. The framework has 2 phases: 1) Offline, a VRED model $f$~\cite{wang2019vred} is trained to predict purely kinematic trajectories $f(h_{0:t}, \delta) = h_{t+1:T}$,  2) Online, trajectory optimization techniques are used to adapt to environmental objectives. This is done by changing additional controls~$\delta$ that are added to the VRED architecture. Possible constraints for trajectory optimization are goalset constraints, collision constraints, or joint objectives between a human and a robot agent.

\subsection{Affordances for Motion Prediction}
Affordances model the idea of the existence of an intuitive and perceptual representation of the possibilities in an environment. It stems its roots from psychology~\cite{gibson2} and is also used in robotics~\cite{koppula2016}.
\revision{In this application of the dataset,} a neural network is used to learn and encode affordances. Details can be found in~\cite{kratzer2020anticipating}.
We focus on pick and place affordances and model them as probability density functions conditioned
on the environment and the kinematic state of
the human.
Figure~\ref{sfig:heatmap} shows an example heatmap predicted by the network for placing a jug on the table. We combine the affordance prediction with the full-body motion prediction from the previous section in Figure~\ref{sfig:fbaffordance}. 
%The motion prediction framework is informed with a sampled position from the affordance heatmap. In Figure~\ref{sfig:fbgt} the ground truth trajectory can be seen. It can be seen that the prediction comes very close to the ground truth, which demonstrates the efficacy of the approach.

\subsection{Hierarchical Motion Prediction}
\label{ssec:HMP}
Human behavior naturally follows different hierarchy levels. While on a higher level there might be goals like setting up a table for 4 persons, the lower level task would be controlling the muscles of the human.

Here, we used maximum entropy inverse reinforcement learning to learn local rewards for the sub-tasks. At the lowest level, we forecast the human base position using the method by Kitani et al.~\cite{kitani2012activity}. \revision{The method uses exponentiated distance functions with different variances as input features as depicted in Figure~\ref{fig:sdf}. The model uses the trajectory over the corresponding
  feature maps to calculate the empirical mean feature count of the demonstrations. The expected state visitation frequencies are calculated and the weights are adapted until the empirical feature counts are matched.}

For the higher level, the model encodes the high-level task objective based on human demonstrations of motion performing the high-level task. \revision{Therefore, the area is divided into specific locations (table, big shelf, and small shelf). Whenever an object enters one of these locations, the corresponding state of that object changes. The path length of the low-level task is used as the action cost in the high-level model.}
An example trajectory can be seen in Table~\ref{tab:hltrajectory}. A compact state representation describes where each object is located. After performing a high-level action like moving a cup from the small shelf to the table, the corresponding location changes.
The model achieved a success rate of 84\% on average for the higher levels on cross-validation tests\revision{, which means that 84\% of the state transition match ground-truth behavior.}

\subsection{Gaze-based Intent Prediction}
\label{ssec:gaze}
One important application to achieve proactive human robot interaction is to predict the human's intention. Examples for intent prediction in our case is to predict where the human is going to place an object or to predict which object the human is going to pick up.

To investigate the advantage of the eye-gaze for intent prediction, we compare different intent prediction models with and without gaze features.

Figure~\ref{fig:pickpred} depicts prediction accuracy for the pick intention for a time span of 3 seconds, with the grasp of the object occuring at the third second. \revision{It shows the following methods: \textit{RNN with gaze} is a recurrent neural network that uses gaze distance and human skeleton positions as input. The \textit{RNN without gaze} is a similar model but without the gaze features. It can be seen that it performs worse than the network with gaze features. The \textit{Gaze Distance Baseline} and \textit{Hand Distance Baseline} are simple baselines that output the object that is closest to the gaze-ray or the hand. The \textit{Gaze Distance Baseline} outperforms other  models that are based on both: human skeleton positions and gaze features.}

An interesting observation is that the gaze baseline performs strongest about half a second before a pick action happens. This is because once the human has planned the grasp, the human then starts to look at other objects or at nearby objects on the table.

The strong performance of the simple gaze baseline shows that leveraging gaze features for intent prediction is very promising.

%%% Local Variables:
%%% mode: latex
%%% TeX-master: "root"
%%% End:

%% file: summary.tex
In this work we presented the \textit{MoGaze} dataset, a novel dataset of human motion data for manipulation tasks. The dataset includes full-body motion, scene data of the objects, and eye-gaze. It is the first to include both: full-body human motion data and eye-gaze data, which was captured using a professional eye-tracking device.

We presented applications that we have performed with the dataset, such as context-aware motion prediction or the use of gaze for intent prediction. The applications demonstrate the strong use of our dataset for the research field of long-term human motion prediction.

For future work, we plan to capture a similar dataset with multiple humans involved at the same time, where humans collaborate on performing a manipulation task.

%%% Local Variables:
%%% mode: latex
%%% TeX-master: "root"
%%% End: